\documentclass[english]{ccdconf}
\usepackage{amsmath}
\usepackage{amsfonts,amssymb}
\usepackage{multirow}
\usepackage[misc]{ifsym} 
\usepackage[numbers,sort&compress]{natbib}

\usepackage{booktabs}
\usepackage{longtable}
\usepackage{float}

\usepackage[colorlinks, linkcolor=red, anchorcolor=blue, citecolor=green]{hyperref}

\begin{document}

\title{MsCGAN: Multi-scale Conditional Generative Adversarial 
Networks for Person Image Generation}
\author{Wei Tang, Gui Li,
      Xinyuan Bao, Teng Li \Letter}

\affiliation{Anhui University, Hefei, 230601, China  \email{liteng@ahu.edu.cn}}

\maketitle

\begin{abstract}
To synthesize high-quality person images with arbitrary poses is challenging. In this paper, we propose a novel Multi-scale Conditional Generative Adversarial Networks (MsCGAN), aiming to convert the input conditional person image to a synthetic image of any given target pose, whose appearance and the texture are consistent with the input image. MsCGAN is a multi-scale adversarial network consisting of two generators and two discriminators. One generator transforms the conditional person image into a coarse image of the target pose globally, and the other is to enhance the detailed quality of the synthetic person image through a local reinforcement network. The outputs of the two generators are then merged into a synthetic, discriminant and high-resolution image. On the other hand, the synthetic image is downsampled to multiple resolutions as the input to multi-scale discriminator networks. The proposed multi-scale generators and discriminators handling different levels of visual features can benefit to synthesizing high-resolution person images with realistic appearance and texture. Experiments are conducted on the Market-1501 and DeepFashion datasets to evaluate the proposed model, and both qualitative and quantitative results demonstrate the superior performance of the proposed MsCGAN.
\end{abstract}


\footnotetext{This work is supported by the National Natural Science Foundation (NSF) of China (No. 61572029) and the Anhui Provincial Natural Science Foundation of China (No. 1908085J25).}

\section{INTRODUCTION}
To generate high-resolution and photo-realistic person image with arbitrary poses is important in multimedia and computer vision for its wide range of applications such as data augmentation in person re-ID\cite{Zheng2017Unlabeled}, person image editing or inpainting\cite{Si_2018_CVPR} and video forecasting\cite{Walker2017The}. Various methods have been proposed to handle this problem, such as VAE\cite{kingma2013auto}, ARMs\cite{oord2016pixel}, and GANs\cite{goodfellow2014generative}. Owing to impressive performances of GANs, recent works on person image generation\cite{Si_2018_CVPR, ma2017pose, siarohin2018deformable, pumarola2018unsupervised, ma2018disentangled}and high-resolution image synthesis\cite{wang2018high, ledig2017photo, brock2018large} mostly focus on it.

From the perspective of the prior knowledge, existing person image generation methods based on GANs can be categorized into two groups. The first is the global pose-guided strategy \cite{ma2017pose, siarohin2018deformable, pumarola2018unsupervised, zhu2019progressive} and the second is the segmentation-guided multi-module strategy \cite{ma2018disentangled, song2019unsupervised}. The former is to synthesize the target person image through a global model by inputting the conditional person image and the human pose estimation simultaneously. The latter is to parse the conditional person image into foreground, background and pose information, then synthesize the target image using multi-module models. Although the pose-guided person image generation methods can be accurate regarding target human poses, they usually neglect the detailed appearance and the texture information of the conditional person images. Segmentation-guided multi-module person image generation techniques can preserve appearance and texture features of the conditional person image more completely. However, most multi-module models are complex and difficult to train. Moreover, the pose accuracy of the resulted image by these models is still far from expectation. In many cases, above-mentioned methods are still problematic to produce person image with precise pose and preserve appearance details simultaneously. 

\begin{figure}[!t]
\setlength{\abovecaptionskip}{-5pt}
\setlength{\belowcaptionskip}{-5pt}
  \centering
   \includegraphics[width=88mm]{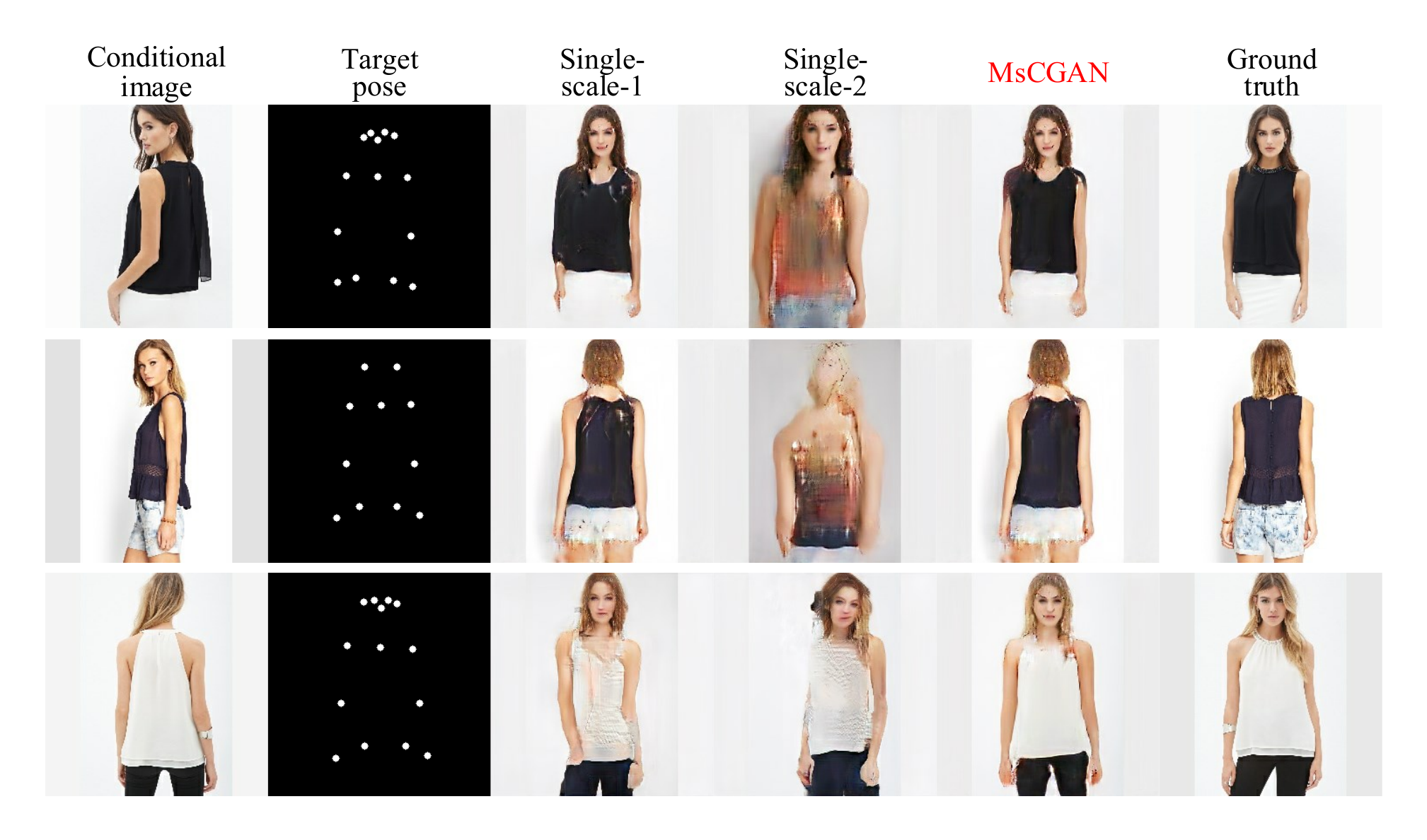}\\
  \caption{Some synthetic samples by the proposed MsCGAN on DeepFashion dataset.}
  \label{fig:fig1}
\end{figure}

In this paper, we propose a novel Multi-scale Conditional Generative Adversarial Networks (MsCGAN) for person image generation. In Figure 1, the inputs are the conditional image image and the target pose. Our goal is to synthesize the person image with the target pose, whose appearance and the texture information are also consistent with the conditional images. MsCGAN is based on pose guidance so that it can synthesize the person images with high pose accuracy. Meanwhile, MsCGAN contains two strategies to ensure the visual quality of the generated images to be consistent with the conditional image. One of them is using the global-to-local generators, which generate a coarse image of the specific pose globally, and refine the coarse image locally. The other is that multi-scale discriminators are adopted to discriminate the generated image and its downsampled images respectively, which aim to handle the visual features on multiple levels.

Compared with existing methods, our proposed model has several advantages:1) Joint the global generation and the local refinement can model both the accuracy and quality of the synthetic image simultaneously. 2) Images with different resolutions contain different levels of visual features, so the proposed multi-scale discriminators can increase the receptive field of the discriminator. 3) The combination of the global-to-local generators and multi-scale discriminators ensure the synthetic person images have the target pose and more detailed appearance features than existing methods. 

Some examples are shown in Figure 1. The proposed model is evaluated on the dataset Market-1501\cite{Zheng2015Scalable}, as well as the fashion dataset DeepFashion \cite{Liu2016DeepFashion}. Moreover, we compare our results to the results of the state-of-the-art methods to demonstrate effectiveness. The main contributions of our work are summarized as follows:
\begin{itemize}
\item A novel MsCGAN model are proposed for person pose image generation, which is able to handle more detailed appearance and texture information while synthesizing high-quality and realistic person images with arbitrary poses. 
\item The proposed global-to-local generating strategy, as well as multi-scale discriminators, can be widely used in GANs based image or video synthesis. 
\item The proposed MsCGAN is a two-stage model, which is more compact than segmentation-guided multi-module methods, and it achieves much superior performance than existing methods on the benchmark Market-1501 and DeepFashion datasets.
\end{itemize}

The rest of the paper is organized as follows. Section 2 introduces the details of the proposed MsCGAN. In Section 3, we report and analyze extensive experimental results. Finally, we conclude the paper with future work in Section 4.

\section{METHODOLOGY} 
Table 1 lists the key notations for the proposed MsCGAN which will be used in the following parts. The architecture of MsCGAN is shown in Figure 2. In summary, the global-to-local generators consist of two generators that integrate the coarse image and enhance the refined images respectively. To obtain high-level features and the texture information of the conditional person image, we apply the multi-scale discriminators discriminate images with different resolutions. Overall, MsCGAN contains two stages that will be detailed respectively as follows:

\subsection{Stage-\uppercase\expandafter{\romannumeral1}: Global Generation}  	

\begin{table}	
  \centering
  \caption{List of key notations}
  \label{tab1}
  \begin{tabular}{l|l}
    \hhline
    \textbf{Notation}	&	\textbf{Description} \\ 	\hline
		\emph{x}	&	the conditional person image \\ 	\hline
		\emph{y}	&	the target person image \\ 	\hline
		\emph{p}	&	the pose of target person image \\	\hline
		$ \hat{y}_{1} $		&	the result of stage-\uppercase\expandafter{\romannumeral1} \\	\hline
		$ \hat{y}_{2} $		&	the result of stage-\uppercase			 \expandafter{\romannumeral2} \\	\hline
		$\hat{y}$	&	the final synthetic image \\	\hline
		$ x_{d} $	&	the image after downsampling $\emph{x}$ \\	\hline
		$ y_{d}	$	&	the image after downsampling \emph{y }\\	\hline
		$\hat{y}_{d}$	&	the image after downsampling $\hat{y}$ \\	\hline
		\textcircled{+}	&	concatenation \\	\hline
		\textcircled{x}	&	weighted sum \\
	  \hhline
  \end{tabular}
\end{table}

At stage-\uppercase\expandafter{\romannumeral1}, we input the conditional person image (\emph{x}) and the human pose estimation of the target person image (\emph{p}) to synthesize a global structure image ($ \hat{y}_1 $) containing the pose of the target person image (\emph{y}).

\subsubsection{Human Pose Estimation}	
Here we use the Part Affinity Fields (PAFs)\cite{cao2017realtime} to obtain a highly accurate map of the human pose estimation. In our work, the PAFs estimate the human pose in the conditional person image and generates the coordinates of eighteen 2D keypoints in real-time. Then we encode the eighteen 2D keypoints into eighteen heatmaps to express the human pose more intuitively.

\subsubsection{Generator $G_1$: Full Person Image Generator}	
Generator $ G_1 $ adopts a variant of the U-Net network  \cite{Ronneberger2015U}  and the residual blocks \cite{He_2016_CVPR}. The input of $ G_1 $ is a concatenation of the conditional person image (\emph{x}) and the map of human pose estimation generated by the target person image (\emph{p}), while the output of $ G_1 $ is a coarse image ($ \hat{y}_1 $) with the pose of the target person image. The poses of the \emph{x} and the ($ \hat{y}_1 $) are quite different, but they are closely related to the appearance feature of the person. To change the pose of the person and preserve the detailed appearance of the person simultaneously, we employ skip connection  between the symmetric layers of encoder and decoder in the variant of U-Net network. Since the generated image using $L_2$ distance is more blurred than that using $L_1$  distance in generator\cite{isola2017image}, we use $L_1$  distance as its basic loss function.

\begin{figure*}
	\setlength{\abovecaptionskip}{-0.2cm}
  \centering
  \includegraphics[width=170mm]{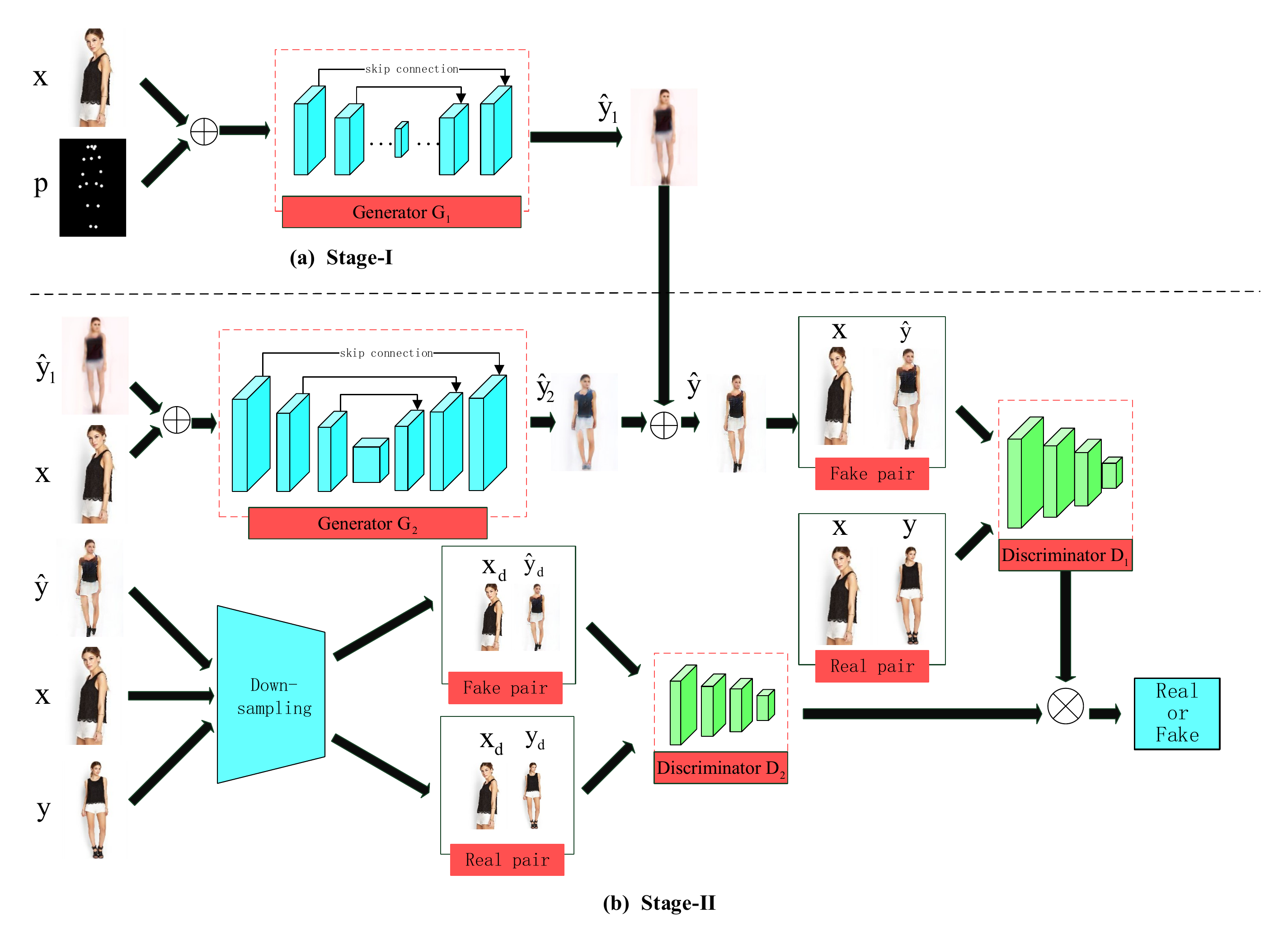}\\
  \caption{The architecture of the proposed MsCGAN.}
  \label{fig:fig2}
\end{figure*}

\subsubsection{Compensation for the Impact of Background}	
The generators originally have no idea what the background of generates image should be like, which would cause the results unrealistic. Furthermore, when the input image is unsharp, the person and the background are prone to jumble. In order to mitigate the impact of background, we introduce the background loss that uses the pose mask of the target person image ($P_m$) to separate the person when calculating $L_1$ loss. In the $P_m$, $1$ indicates the person and $0$ indicates the background.

\begin{equation}	
\label{eq1}
\begin{split}
\mathcal L _{Stage\_1}~=~\mathcal L_{L1\_1}+\lambda_{bg\_1}\mathcal L_{bg\_1}~~~~~~~~~~~~~~~~~~~~~~~~~~~~~~~~~~~~~~~~~\\
~=~\parallel y-G_1(x,p)\parallel_1+\lambda_{bg\_1}\parallel (y-G_1(x,p))\odot p_m\parallel_1,
\end{split}
\end{equation}
here $ \|~y- G_1(x,p)~\|_1$ and $\|~y- G_1(x,p) \odot P_m~\|_1 $ indicate $L_1$ loss and background loss, $\lambda_{bg\_1}$ is a hyper-parameter that indicates the weight of background loss, and $\odot$ denotes the pixels-wise multiplication.

\subsection{Stage-\uppercase\expandafter{\romannumeral2}: Local Reinforcement}  
At stage-\uppercase\expandafter{\romannumeral2}, we correct the detailed feature of $ \hat{y}_1 $ and discriminate against the synthetic image. We design a local reinforcement generator $ G_2 $ to generate an appearance difference map ($ \hat{y}_2 $) between \emph{x} and $ \hat{y}_1 $. Then, $ \hat{y}_1 $ and $ \hat{y}_2 $ are merged into the final synthetic image ($ \hat{y} $). Multi-scale discriminators discriminate against the synthetic images with different resolutions. The conditional DCGAN-like network are adopted as our base model.

\subsubsection{Generator $G_2$: Local Reinforcement Generator}	
Local reinforcement generator $G_2$ is a network inspired by conditional DCGAN and U-Net architectures. Specifically, the encoder of $G_2$ is a variant of the conditional DCGAN network \cite{Radford2015Unsupervised} and is symmetric to the decoder of $G_2$. The inputs of $G_2$ are the conditional person image (\emph{x}) and the stage-\uppercase\expandafter{\romannumeral1} generation result ($ \hat{y}_1 $), while the output of $G_2$ is an appearance difference map ($ \hat{y}_2 $) between \emph{x} and $ \hat{y}_1 $. The fully connected layer destroys the spatial structure of the image while compressing it. The DCGAN network is a full-convolution architecture so that the details of the image can be preserved more completely.

\subsubsection{Multi-scale Discriminators} 
In order to generate a high-resolution image, it is necessary to increase the receptive field of the discriminator. We adopt multi-scale discriminators which consist of two discriminators ($D_1$ and $D_2$) that discriminate different resolution versions of the synthetic image. We downsample the conditional person image (\emph{x}), the final synthetic image ($ \hat{y} $) and the target person image (\emph{y}) by a factor of $2$. The discriminators $D_1$ and $D_2$ are trained to distinguish real and synthesized images at the two different scales, respectively.

\begin{figure*}
\setlength{\abovecaptionskip}{-0.2cm}
  \centering
  \includegraphics[width=170mm]{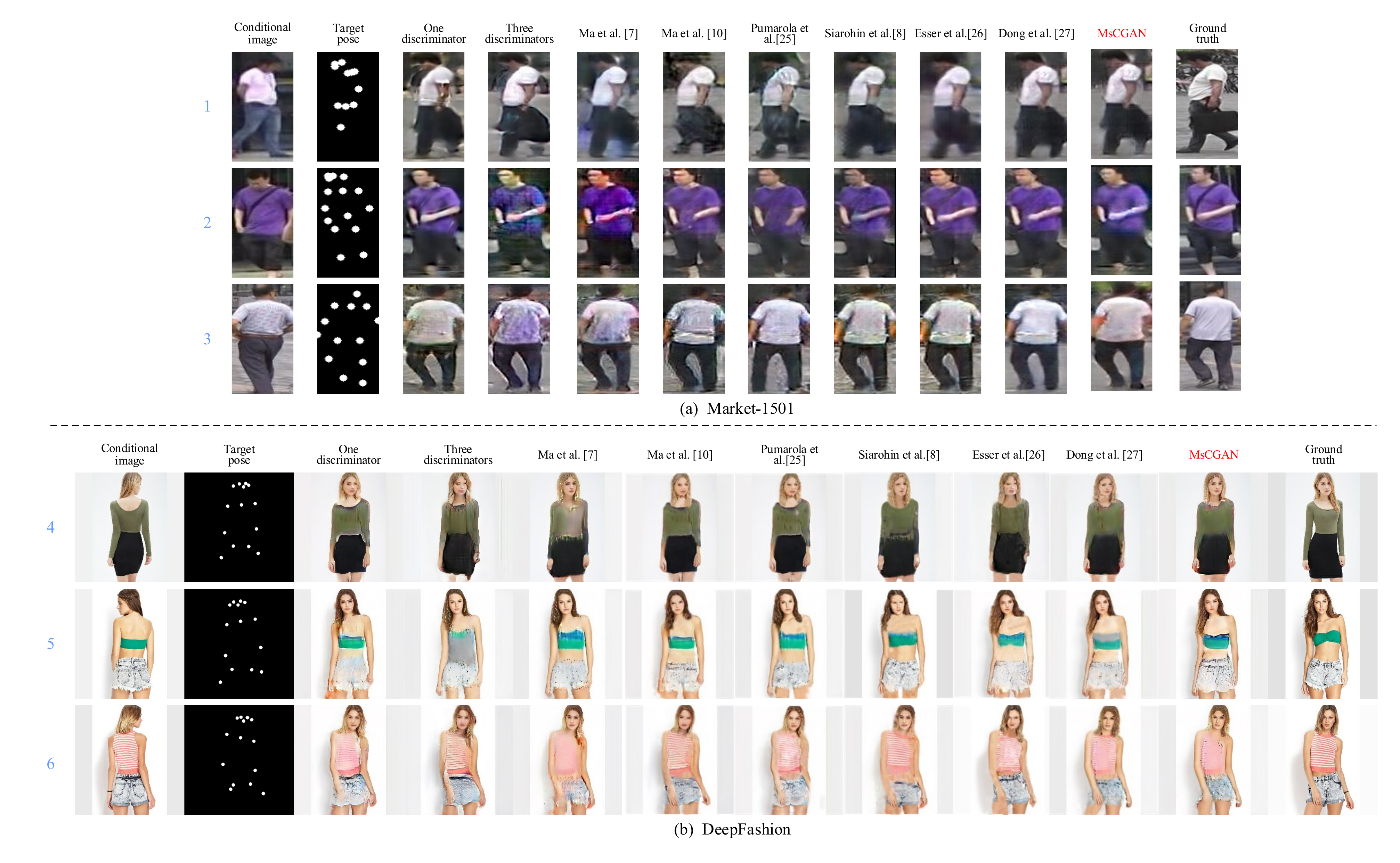}\\
  \caption{Example of results on Market-1501 dataset and DeepFashion dataset.}
  \label{fig:fig3}
\end{figure*}

\subsection{Overall Loss Function for Stage-\uppercase\expandafter{\romannumeral2}}  
At stage-\uppercase\expandafter{\romannumeral2}, we use conditional GAN as our basic loss function and add $L_1$ loss, background loss which introduced in stage-\uppercase\expandafter{\romannumeral1} as extra supervisory signals. The loss function of the conditional GAN is formulated as:
\begin{equation}	
\begin{split}
\mathcal L _{cGAN}(G,D)~= ~~~~~~~~~~~~~~~~~~~~~~~~~~~~~~~~~~~~~~~~~~~~~~~~~~~~~~~
\\ ~\mathbb{E}_{x,y}[logD(x,y)]~+~\mathbb{E}_{x,z}[log(1-D(x,G(x,z)))],
\end{split}
\end{equation}
where \emph{z} refers to $ G_1 (x,p) $. \emph{G} tries to minimize this objective against an adversarial \emph{D} which tries to maximize it. Therefore, the loss function of the conditional GAN can be optimized as follows:
\begin{equation}	
\min \limits_{G_2}\max \limits_{D_1,D_2}(\lambda_{D_1} \mathcal L _{cGAN}(G_2,D_1)+\lambda_{D_2} \mathcal L _{cGAN}(G_2,D_2)),
\end{equation}
where $ \lambda_{D1} $ and $ \lambda_{D2} $ control the importance of the two terms, respectively.

Thus, our full loss combining conditional GAN loss, $L_1$ loss and background loss for Stage-\uppercase\expandafter{\romannumeral2} is formulated as:	
\begin{equation}	
\begin{split}
\mathcal L _{Stage\_2}~=
~\min \limits_{G_2} (\max \limits_{D_1,D_2}(\lambda_{D_1} \mathcal L _{cGAN}(G_2,D_1)+~~~~~~~~~~~~~~~\\
\lambda_{D_2} \mathcal L _{cGAN}(G_2,D_2)) + \lambda_{L1\_2}\parallel y-G_2(x,G_1 (x,p))\parallel_1 \\   	+~\lambda_{bg\_2}\parallel y-G_2(x,G_1 (x,p))\odot p_m \parallel_1),~~~~~~~~~~~~~~~~~~~~~~
\end{split}
\end{equation}
where $\parallel y-G_2(x,G_1 (x,p))\parallel_1$ is $L_1$ loss and $\parallel y-G_2(x,G_1 (x,p))\odot p_m \parallel_1$ is background loss respectively, $\lambda_{L1\_2}$ and $ \lambda_{bg\_2} $ are the weights of them.

\section{EXPERIMENTS} 	
This section demonstrates the effectiveness of the proposed MsCGAN. Firstly, the datasets and implementation details are introduced. Then, we show the qualitative and quantitative evaluations of the proposed MsCGAN.	

\subsection{Datasets and Implementation Details}	
We conduct evaluations on two public datasets: Market-1501 dataset \cite{Zheng2015Scalable} and DeepFashion dataset \cite{Liu2016DeepFashion}. Our MsCGAN is implemented with Tensorflow \cite{Abadi2016TensorFlow} and we apply the Adam optimizer \cite{Kingma2014Adam} with adapted hyper-parameters. As suggested in \cite{goodfellow2014generative}, we improve the gradient by training to maximize rather than minimize in the implementation.

\subsection{Baselines}	
To demonstrate the proposed MsCGAN can achieve superior results, we compare our method with several state-of-the-art works: \cite{ma2017pose, ma2018disentangled, Pumarola_2018_CVPR, siarohin2018deformable, esser2018variational, NIPS2018_7329}. Moreover, to demonstrate the proposed network structure proposed in Section 2 is optimal, we evaluate qualitative and quantitative results of three settings: One discriminator, Two discriminators (MsCGAN) and Three discriminators. All of them have the same global-to-local generators while the number of discriminators are different. Note that we use the same generators, the same datasets and the same hyper-parameters to train and test all baselines mentioned above.

\begin{table*}
    \centering
    \caption{Quantitative results on two public datasets, higher scores are better.}
    \label{table2}
    \renewcommand\arraystretch{1.3}
    \begin{tabular}{p{3.3cm} | p{1.5 cm} p{1.5 cm} | p{1.5 cm} p{1.5 cm} p{1.5 cm} p{1.5 cm}}
        \hhline 
        \multirow{2}{*}{Model} & \multicolumn{2} {c|} {DeepFashion} & \multicolumn{4}{c}{Market-1501} \\
         	& SSIM & IS & SSIM & IS & mask-SSIM & mask-IS \\  
        \hline
        One discriminator &	0.737	& 3.212 & 0.248	& 3.167 & 0.791 & 3.246 \\  \hline 
        \textbf{MsCGAN} &	0.725	& 3.335 & 0.219	& \textbf{3.588} & 0.764 & \textbf{3.741} \\  \hline 
        Three discriminators &	0.691	& 3.284 & 0.216	& 3.459 & 0.778 & 3.558 \\  \hline 
        Ma et al.\cite{ma2017pose} &	0.762	& 3.090 & 0.253	& 3.460 & 0.792 & 3.435 \\  \hline 
        Ma et al.\cite{ma2018disentangled} &	0.614	& 3.228 & 0.099	& 3.483 & 0.614 & 3.491 \\  \hline 
        Pumarola et al.\cite{Pumarola_2018_CVPR} &	0.747	& 2.97 & ---	& --- & --- & --- \\  \hline 
        Siarohin et al.\cite{siarohin2018deformable} & 0.756 & \textbf{3.439} & 0.290 & 3.185 & \textbf{0.805} & 3.502 \\  \hline 
        Esser et al.\cite{esser2018variational} & 0.786 & 3.087 & 0.353 & 3.214 & --- & --- \\  \hline 
        Dong et al.\cite{NIPS2018_7329} & \textbf{0.793} & 3.314 & \textbf{0.356} & 3.409& --- & --- \\ 
        \hhline	
    \end{tabular}
\end{table*}

\subsection{Results and Discussions} 
\subsubsection{Qualitative Results}  
We apply global-to-local generators and multi-scale discriminators consisting of two discriminators. To illustrate the difference between multi-scale discriminators, we perform different number of discriminators for qualitative comparisons. Moreover, we compare our results with six different methods \cite{ma2017pose, ma2018disentangled, Pumarola_2018_CVPR, siarohin2018deformable, esser2018variational, NIPS2018_7329}. The results for different models are shown in Figure 3. It is clear that MsCGAN can generate more photo-realistic results than other methods. The major reason is that our MsCGAN plays more attention on the appearance and the texture information of the conditional person image.

\subsubsection{Quantitative Results}
We adopt the evaluation protocol from previous person image generation works \cite{ma2017pose, siarohin2018deformable, pumarola2018unsupervised, ma2018disentangled, Johnson2016Perceptual, Shi2016Real}. We use Structural Similarity Index Measure (SSIM) \cite{Zhou2004Image} and Inception Score (IS) \cite{Salimans2016Improved} and  their corresponding masked versions mask-SSIM and mask-IS \cite{ma2017pose}. It is objective to evaluate the IS on a large enough number of samples \cite{Salimans2016Improved}. Quantitative results are shown in Table 2.

From Table 2, it is clear that MsCGAN achieves the highest or high score in IS and mask-IS criteria. There is still a gap between MsCGAN and the state-of-the-art approach in SSIM and mask-SSIM criteria.

\subsubsection{Failure Cases}
Although MsCGAN can synthesize more photo-realistic results than other methods, it still has failure cases. We select a few typical failure cases that are shown in Figure 4. In row 1, we can see that objects in the conditional person image have a great influence on the human pose estimation so the proposed MsCGAN synthesizes poor-quality result. From the result in row 2, the occlusion of the person in the conditional person image causes the appearance of the synthetic image to be disorganized.

In row 3, we can find that MsCGAN causes an error of gender which transforms a female into a male. In row 4, the body in the target person image does not appear in the conditional person image, resulting in an incomplete body of the generated image.
There are two main reasons why image generation fails. The first one is that the conditional person image is too blurred and there are else objects in it, which affects the detection of keypoints on the human body. The second one is that the conditional person image is occluded, which affects the appearance features and texture information of the synthetic image.

\begin{figure}
\setlength{\abovecaptionskip}{-5pt}
\setlength{\belowcaptionskip}{-5pt}
  \centering
   \includegraphics[width=88mm]{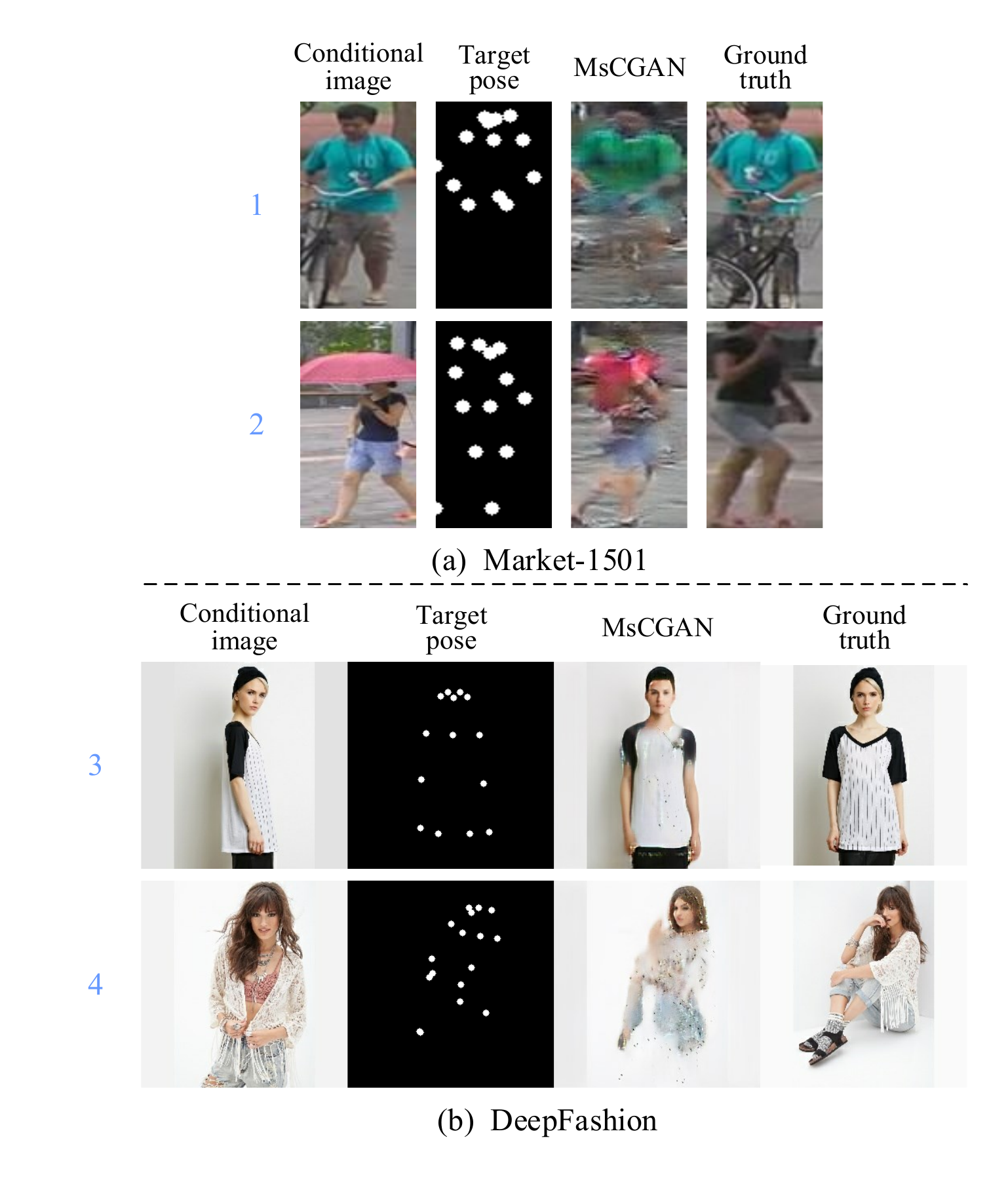}\\
  \caption{A few typical failure cases.}
  \label{fig:fig4}
\end{figure}

\section{CONCLUSION} 
In this paper, we propose a novel multi-scale conditional generative adversarial networks (MsCGAN) for generating high-quality and photo-realistic person image with arbitrary poses. The proposed MsCGAN can simultaneously model target pose as well as the appearance of the conditional person image in a unified and principled way. In the experiments, the qualitative and quantitative results clearly demonstrate the effectiveness of the proposed model. In the future, we will evaluate our model on more practical datasets and investigate more applications.

%
%
\balance


\section*{Appendix}
\label{sec:Appendix}
We list the detailed network architectures of MsCGAN in the following tables. The key notations for the abbreviated words which represent the network structures are shown in Table~\ref{table3}. Since network layers of MsCGAN depend on the size of the input images, MsCGAN is a novel model with variational network layers.

\begin{table}[H]
    \setlength{\abovecaptionskip}{0cm} 
    \centering
    \caption{List of key notations for the abbreviated words.}
    \label{table3}
    \begin{tabular}{l|l}
        \hhline
        \textbf{Notation}	&	\textbf{Description} \\
        \hline
        K	&	Kernel layer \\
        \hline
        S	&	Stride \\
        \hline
        P	&	Padding \\
        \hline
        Act	&	Activation function \\
        \hline
        OM	&	Output channels for Market-1501 dataset\\
        \hline
        OD	&	Output channels for DeepFashion dataset\\
        \hline
        Con	&	Convolutional layer  \\
        \hline
        Dec	&	Deconvolutional layer  \\
        \hline
        ConB	&	Convolutional block \\
        \hline
        ResB	&   Residual block \\
        \hline
        FC	&   Fully connected layer \\
        \hline
        BN	&   BatchNorm \\
        \hhline
    \end{tabular}  
\end{table}

\subsection*{A~~~~The Network Architectures of Generator $G_1$}
The network architectures of $G_1$ are shown in Table~\ref{table4} and Table~\ref{table5}. \emph{Conv2}, \emph{ResB5}, \emph{Dec1} and \emph{ResB5} are only for DeepFashion dataset. The rest of the network structures are applicable to both Market-1501 dataset and DeepFashion dataset. In the decoder, the last layer of the first three residual blocks is a deconvolutional layer which is used to enlarge the generated images.

\begin{table}[H]
    \setlength{\abovetopsep}{0.5ex}	
    \setlength{\belowrulesep}{0pt}
    \setlength{\aboverulesep}{0pt}
   
    \setlength{\abovecaptionskip}{0cm} 
    \centering
    \caption{The encoder of $G_1$.}
    \label{table4}
    \begin{tabular}{ccccccc}
        \hhline
        Layer & K & S & P & Act & OM & OD\\
        \midrule
        Con1 &	3*3	& 1 & 1	& ReLU & 128 & 128 \\	\cmidrule{1-7}
        \multirow{3}{*}{ResB1} &	3*3	& 1 & 1	& ReLU & 128 & 128 \\
        & 3*3	& 1 & 1	& ReLU & 128 & 128 \\
        & 3*3	& 2 & 1	& ReLU & 256  & 256\\	\cmidrule{1-7}
        \multirow{3}{*}{ResB2} &	3*3	& 1 & 1	& ReLU & 256 & 256 \\
        & 3*3	& 1 & 1	& ReLU & 256  & 256\\
        & 3*3	& 2 & 1	& ReLU & 512  & 512\\	\cmidrule{1-7}
        \multirow{3}{*}{ResB3} &	3*3	& 1 & 1	& ReLU & 512  & 512\\
        & 3*3	& 1 & 1	& ReLU & 512 & 512 \\	
        & 3*3	& 2 & 1	& ReLU & 1024  & 1024\\	\cmidrule{1-7}
        \multirow{2}{*}{ResB4} &	3*3	& 1 & 1	& ReLU & 1024 & 1024 \\
        & 3*3	& 1 & 1	& ReLU & 1024  & 1024\\	\cmidrule{1-7}
        \emph{\textbf{Con2}} &	3*3	& 2 & 1	& ReLU & ---  & 2048\\	\cmidrule{1-7}
        \multirow{2}{*}{\emph{\textbf{ResB5}}} &	3*3	& 1 & 1	& ReLU & ---  & 2048\\
        	& 3*3	& 1 & 1	& ReLU & ---  & 2048\\	\cmidrule{1-7}
        	
         FC &--- &--- &--- &--- & 64 dims & 64 dims\\
			\hhline
      \end{tabular}
\end{table}

\begin{table}[H]
	\setlength{\abovetopsep}{0.5ex}	
	\setlength{\belowrulesep}{0pt}
	\setlength{\aboverulesep}{0pt}
	\setlength{\abovecaptionskip}{0cm} 
	\centering
	\caption{The decoder of $G_1$.}
	\label{table5}
	\begin{tabular}{ccccccc}
		\hhline
		Layer & K & S & P & Act & OM & OD \\
		\midrule
		FC &--- &--- &--- &--- & 4096 dims & 4096 dims \\ 	\cmidrule{1-7}
		\multirow{3}{*}{ResB1} &	3*3	& 1 & 1	& ReLU & 1024& 2048  \\
			& 3*3	& 1 & 1	& ReLU & 1024 & 2048 \\
			& 3*3 & 2 & 1 & ReLU & 512 & 1024 \\		\cmidrule{1-7}
		\multirow{3}{*}{ResB2} &	3*3	& 1 & 1	& ReLU & 512 &1024  \\
			& 3*3	& 1 & 1	& ReLU & 512 & 1024  \\
			& 3*3 & 2 & 1 & ReLU & 256 & 512 \\		\cmidrule{1-7}
		\multirow{3}{*}{ResB3} &	3*3	& 1 & 1	& ReLU & 256  & 512 \\
			& 3*3	& 1 & 1	& ReLU & 256  & 512 \\
			& 3*3 & 2 & 1  & ReLU & 128 & 256 \\		\cmidrule{1-7}
		\multirow{2}{*}{ResB4} &	3*3	& 1 & 1	& ReLU & 128 & 256  \\
			& 3*3	& 1 & 1	& ReLU & 128 & 256   \\		\cmidrule{1-7}	
		\textit{\textbf{Dec1}} & 3*3 & 2 & 1 & ReLU & ---  & 128 \\	\cmidrule{1-7}
		\multirow{2}{*}{\textit{\textbf{ResB5}}} &	3*3	& 1 & 1	& ReLU & ---  & 128  \\
			& 3*3	& 1 & 1	& ReLU & ---  & 128   \\	\cmidrule{1-7}      	
			Con1 & 3*3 & 1 & 1 & ReLU & 3 & 3\\
		\hhline
	\end{tabular}
\end{table}

\subsection*{B~~~~The Network Architectures of Generator $G_2$}
The network architectures of $G_2$ are shown in Table~\ref{table6} and Table~\ref{table7}. \emph{Con2}, \emph{ConB4}, \emph{Dec1} and \emph{ConB4} are only for DeepFashion dataset. Similarly to $G_1$, the rest of the network structures are applicable to both two datasets and the last layer of the first two convolutional blocks is a deconvolutional layer.

\begin{table}[H]
    \setlength{\abovetopsep}{0.5ex}	
    \setlength{\belowrulesep}{0pt}
    \setlength{\aboverulesep}{0pt}
    
    \setlength{\abovecaptionskip}{0cm} 
    \centering
    \caption{The encoder of $G_2$.}
    \label{table6}
    \begin{tabular}{ccccccc}
        \hhline
        Layer & K & S & P & Act & OM & OD \\
        \midrule
        
        Con1 &	3*3	& 1 & 1	& ReLU & 128 & 128 \\
        \cmidrule{1-7}
        \multirow{3}{*}{ConB1} &	3*3	& 1 & 1	& ReLU & 128 & 128 \\
        & 3*3	& 1 & 1	& ReLU & 128  & 128\\
        & 4*4	& 2 & 1	& ReLU & 256  & 256\\
        \cmidrule{1-7}
        \multirow{3}{*}{ConB2} &	3*3	& 1 & 1	& ReLU & 256 & 256 \\
        & 3*3	& 1 & 1	& ReLU & 256  & 256\\
        & 4*4	& 2 & 1	& ReLU & 512  & 512\\
        \cmidrule{1-7}
        \multirow{2}{*}{ConB3} &	3*3	& 1 & 1	& ReLU & 512 & 512 \\
        & 3*3	& 1 & 1	& ReLU & 512   & 512 \\
        \cmidrule{1-7}	
        \textit{\textbf{Con2}}	& 4*4	& 2 & 1	& ReLU & ---  & 1024\\	
        \cmidrule{1-7}
        \multirow{2}{*}{\textit{\textbf{ConB4}}} &	3*3	& 1 & 1	& ReLU & --- & 1024\\
        & 3*3	& 1 & 1	& ReLU & --- & 1024\\
        \hhline
    \end{tabular}
\end{table}

\begin{table}[!htbp]
    \setlength{\abovetopsep}{0.5ex}	
    \setlength{\belowrulesep}{0pt}
    \setlength{\aboverulesep}{0pt}
    
    \setlength{\abovecaptionskip}{0cm} 
    \centering
    \caption{The decoder of $G_2$.}
     \label{table7}
    \begin{tabular}{ccccccc}
        \hhline
         Layer & K & S & P & Act & OM & OD \\
        
        \midrule
        \multirow{3}{*}{ConB1} &	3*3	& 1 & 1	& ReLU & 512 & 1024 \\
        	& 3*3	& 1 & 1	& ReLU & 512  & 1024\\
          & 4*4	& 2 & 1	& ReLU & 256  & 512 \\
        \cmidrule{1-7}
        \multirow{3}{*}{ConB2} &	3*3	& 1 & 1	& ReLU & 256 & 512\\
         	& 3*3	& 1 & 1	& ReLU & 256 & 512\\
           & 4*4	& 2 & 1	& ReLU & 128 & 256 \\
        \cmidrule{1-7}
        \multirow{2}{*}{ConB3} &	3*3	& 1 & 1	& ReLU & 128 & 256\\
        	& 3*3	& 1 & 1	& ReLU & 128 & 256\\
        \cmidrule{1-7}	
        \textit{\textbf{Dec1}} & 4*4	& 2 & 1	& ReLU & --- & 128 \\
        \cmidrule{1-7}
        \multirow{2}{*}{\textit{\textbf{ConB4}}} &	3*3	& 1 & 1	& ReLU & --- & 128\\
        	& 3*3	& 1 & 1	& ReLU & --- & 128\\
        \cmidrule{1-7}
         Con1 & 3*3	& 1 & 1	& ReLU & 3 & 3\\
        \hhline
    \end{tabular}
\end{table}

\subsection*{C~~~~The Network Architectures of Discriminators }
The network architectures of Discriminators are shown in Table~\ref{table8} and \emph{Con5} is just for DeepFashion dataset. The first four convolutional layers are applicable to both Market-1501 dataset and DeepFashion dataset.
    
\begin{table}[!htbp]
    \setlength{\abovetopsep}{0.5ex}	
    \setlength{\belowrulesep}{0pt}
    \setlength{\aboverulesep}{0pt}
    
    \setlength{\abovecaptionskip}{0cm} 
    \centering
    \caption{The network architectures of discriminators}
    \label{table8}
    \begin{tabular}{ccccccc}
        \hhline
        Layer & K & S & P & BN & Act & O \\
        \midrule
        Con1 & 4*4 & 2 & 1 & No & LeakyReLU & 128 \\
        Con2 & 4*4 & 2 & 1 & Yes & LeakyReLU & 256 \\
        Con3 & 4*4 & 2 & 1 & Yes & LeakyReLU & 512 \\
        Con4 & 4*4 & 2 & 1 & Yes & LeakyReLU & 1024 \\
        \textit{\textbf{Con5}} & 4*4 & 2 & 1 & Yes & LeakyReLU & 2048 \\
        \hhline
    \end{tabular}
    
\end{table}

\bibliographystyle{IEEEtran}
\setlength{\bibsep}{0.0ex}    
\bibliography{ref}

%
%
%

\end{document}